\documentclass{article}

\usepackage{microtype}
\usepackage{graphicx}
\usepackage{booktabs} % for professional tables

\usepackage{subfig}
\usepackage{xcolor}
\usepackage{hyperref}
\usepackage{enumitem}
\usepackage{multirow}
\usepackage[accepted]{icml2023}

\usepackage{amsmath}
\usepackage{amssymb}
\usepackage{mathtools}
\usepackage{amsthm}

\usepackage[capitalize,noabbrev]{cleveref}

\usepackage{mathtools}
\usepackage{amssymb}
\usepackage{amsthm}
\usepackage{color}
\newcommand{\cut}[1]{{}}
\newcommand{\tA}{{\tilde{\vA}}}

\newcommand{\tL}{{\tilde{\vL}}}

\newcommand{\Xin}{{\vX_{\text{in}}}}

\newcommand{\tin}{{\text{in}}}

\newcommand{\grad}{{\text{grad}}}
\newcommand{\fea}{{\text{fea}}}

\usepackage{multirow}
\newcommand{\mr}[2]{\multirow{#1}{*}{#2}}
\newcommand{\mc}[3]{\multicolumn{#1}{#2}{#3}}

\newcommand{\vA}{{\mathbf{A}}}
\newcommand{\vB}{{\mathbf{B}}}

\newcommand{\vD}{{\mathbf{D}}}

\newcommand{\vG}{{\mathbf{G}}}

\newcommand{\vI}{{\mathbf{I}}}
\newcommand{\vJ}{{\mathbf{J}}}

\newcommand{\vL}{{\mathbf{L}}}
\newcommand{\vM}{{\mathbf{M}}}

\newcommand{\vW}{{\mathbf{W}}}
\newcommand{\vX}{{\mathbf{X}}}
\newcommand{\vY}{{\mathbf{Y}}}

\newcommand{\cE}{{\mathcal{E}}}

\newcommand{\cG}{{\mathcal{G}}}

\newcommand{\cL}{{\mathcal{L}}}

\newcommand{\cO}{{\mathcal{O}}}

\newcommand{\cV}{{\mathcal{V}}}

\newcommand{\RR}{\mathbb{R}}

\newcommand{\vzero}{\mathbf{0}}

\newcommand{\tr}{{\mathrm{tr}}} % trace
\makeatletter
\let\@@span\span
\def\sp@n{\@@span\omit\advance\@multicnt\m@ne}
\makeatother

\newcommand{\bc}{\begin{center}}
\newcommand{\ec}{\end{center}}

\newcommand{\bdm}{\begin{displaymath}}
\newcommand{\edm}{\end{displaymath}}

\newcommand{\beq}{\begin{equation}}
\newcommand{\eeq}{\end{equation}}

\newcommand{\bfl}{\begin{flushleft}}
\newcommand{\efl}{\end{flushleft}}

\newcommand{\bt}{\begin{tabbing}}
\newcommand{\et}{\end{tabbing}}

\newcommand{\beqn}{\begin{align}}
\newcommand{\eeqn}{\end{align}}

\newcommand{\beqs}{\begin{align*}} % no equation numbers
\newcommand{\eeqs}{\end{align*}}  % no equation numbers

\newtheorem{theorem}{Theorem}

\newtheorem{remark}{Remark}
\usepackage[textsize=tiny]{todonotes}

\icmltitlerunning{LazyGNN: Large-Scale Graph Neural Networks via Lazy Propagation}

\begin{document}

\twocolumn[

\icmltitle{LazyGNN: Large-Scale Graph Neural Networks via Lazy Propagation}




\begin{icmlauthorlist}
\icmlauthor{Rui Xue}{first}
\icmlauthor{Haoyu Han}{second}
\icmlauthor{MohamadAli Torkamani}{third}
\icmlauthor{Jian Pei}{fourth}
% \icmlauthor{Xiaorui Liu}{last}
\icmlauthor{Xiaorui Liu}{first}
\end{icmlauthorlist}

% \icmlaffiliation{first}{Department of Electrical and Computer Engineering, North Carolina State University, Raleigh, US}
% \icmlaffiliation{first}{Department of Electrical and Computer Engineering, North Carolina State University, Raleigh, US}
\icmlaffiliation{first}{North Carolina State University, Raleigh, US}
% \icmlaffiliation{second}{Department of Computer Science, Michigan State University, East Lansing, US}
\icmlaffiliation{second}{Michigan State University, East Lansing, US}
% \icmlaffiliation{third}{Amazon, US}
\icmlaffiliation{third}{Amazon, US (this work does not relate to the author's position at Amazon)}
% \icmlaffiliation{fourth}{Department of Computer Science, Duke University, Durham, US}
\icmlaffiliation{fourth}
{Duke University, Durham, US}
% \icmlaffiliation{last}{Department of Computer Science, North Carolina State University, Raleigh, US}

\icmlcorrespondingauthor{Xiaorui Liu}{xliu96@ncsu.edu}

\icmlkeywords{Machine Learning, ICML}

\vskip 0.3in
]



\printAffiliationsAndNotice{}  % leave blank if no need to mention equal contribution

\begin{abstract}

Recent works have demonstrated the benefits of capturing long-distance dependency in graphs by deeper graph neural networks (GNNs). But deeper GNNs suffer from the long-lasting scalability challenge due to the neighborhood explosion problem in large-scale graphs. In this work, we propose to capture long-distance dependency in graphs by shallower models instead of deeper models, which leads to a much more efficient model, LazyGNN, for graph representation learning. Moreover, we demonstrate that LazyGNN is compatible with existing scalable approaches (such as sampling methods) for further accelerations through the development of mini-batch LazyGNN. 
Comprehensive experiments demonstrate its superior prediction performance and scalability on large-scale benchmarks. 
The implementation of LazyGNN is available at \url{https://github.com/RXPHD/Lazy_GNN}.

\end{abstract}
% \vspace{-0.2in}
\section{Introduction}
\label{sec:intro}

Graph neural networks (GNNs) have been widely used for representation learning on graph-structured data~\cite{hamilton2020graph, ma2021deep}, and they achieve promising state-of-the-art performance on various general graph learning tasks, such as node classification, link prediction, and graph classification~\cite{kipf2016semi,gasteiger2018combining,velivckovic2017graph,wu2019simplifying} as well as a variety of important applications, such as recommendation systems, social network analysis, and transportation prediction. 
%%%%%%
In particular, recent research in deeper GNNs has generally revealed the performance gains from capturing long-distance relations in graphs by stacking more graph convolution layers or unrolling various fixed point iterations~\cite{gasteiger2018predict, gu2020implicit, liu2020towards, chen2020simple, li2021training, ma2020unified, pan2020_unified, zhu2021interpreting, chen2020graph}. 
However, the recursive feature propagations in deeper GNNs lead to the well-known neighborhood explosion problem since the number of neighbors grows exponentially with the number of feature propagation layers~\cite{hamilton2017inductive, chen2018fastgcn}. 
This causes tremendous scalability challenges for data sampling, computation, memory, parallelism, and end-to-end training when employing GNNs on large-scale graphs.
It greatly limits GNNs' broad applications in large-scale industry-level applications due to limited computation and memory resources~\cite{ying2018graph, shao2022distributed}.

A large body of existing research improves the scalability and efficiency of large-scale GNNs using various innovative designs, such as sampling methods, pre-computing or post-computing methods, and distributed methods. 
Although these approaches mitigate the neighborhood explosion problem, they still face various limitations when they are applied to deeper GNNs. For instance, sampling approaches~\cite{hamilton2017inductive, chen2018fastgcn, Zeng2020GraphSAINT, zou2019layer,fey2021gnnautoscale, yu2022graphfm} usually incur large approximation errors and suffer from performance degradation as demonstrated in large-scale OGB benchmarks or require large additional memory or storage space to store activation values in hidden layers.
Pre-computing or post-computing methods~\cite{wu2019simplifying, rossi2020sign, sun2021scalable, zhang2022graph, bojchevski2020scaling, zhu2005semi, huang2020combining} lose the benefits of end-to-end training and usually suffer from performance loss. Distributed methods~\cite{chiang2019cluster, chai2022distributed, shao2022distributed} distribute large graphs to multiple servers for parallel training, but they either neglect the inter-server edges or suffer from expensive feature communication between servers.

In this work, we take a substantially different and novel perspective and propose to capture long-distance dependency in graphs by shallower GNNs instead of deeper ones. The key intuition comes from the fact that node information will be propagated over the graph many times during the training process so we may only need to propagate information lazily by reusing the propagated information over the training iterations. This intuition leads to the proposed LazyGNN that solves the inherent neighborhood explosion problem by significantly reducing the number of aggregation layers while still capturing long-distance dependency in graphs through lazy propagation. Multiple technical challenges such as the risk of over-smoothing, additional variation due to feature dropout, and back-propagation through historical computation graphs are addressed through innovative designs in LazyGNN. 
Moreover, since LazyGNN is a shallow model, it naturally tackles the limitations that existing scalable approaches suffer from when handling large-scale and deeper GNNs. Therefore, the contribution of LazyGNN is orthogonal and complementary to existing efforts, and various scalable approaches can be used in LazyGNN for further acceleration. We demonstrate this by developing a highly scalable and efficient mini-batch LazyGNN based on sampling methods. 
To the best of our knowledge, LazyGNN is the first GNN model being versatilely friendly to data sampling, computation, memory, parallelism, and end-to-end training but still captures long-distance dependency in graphs.
Our contributions can be summarized as follows:
% \vspace{-0.2in}
\begin{itemize}
\itemsep0em 
    \item We reveal a novel perspective to solve the neighborhood explosion problem by exploiting the computation redundancy in training GNNs;
    
    \item A novel shallow model, LazyGNN, is proposed to capture long-distance dependency in graphs for end-to-end graph representation learning through lazy forward and backward propagations; 
    
    \item We demonstrate that existing scalable approaches can be compatible with LazyGNN by introducing a highly efficient and scalable mini-batch LazyGNN to handle large-scale graphs based on sampling methods.
    
    \item Comprehensive experiments and studies demonstrate that LazyGNN achieves superior prediction performance and efficiency on large-scale graph datasets.
\end{itemize}

% \section{Preliminaries}
\section{Preliminary}
\label{sec:preliminary_section}

\textbf{Notations.} 
A graph is represented by $\cG = (\cV, \cE)$ where $\cV = \{v_1, \dots, v_N\}$ is the set of nodes and $\mathcal{E} = \{e_1, \dots, e_M\}$ is the set of edges. Suppose that each node is associated with a $d$-dimensional feature vector, and the original features for all nodes are denoted as $\vX_\fea \in \RR^{N\times d}$.
The graph structure of $\cG$ can be represented by an adjacency matrix $\vA \in \RR^{N\times N}$, where $\vA_{ij}>0$ when there exists an edge between node $v_i$ and $v_j$, otherwise $\vA_{i,j}=0$. 
The symmetrically normalized graph Laplacian matrix is defined as $\tL=\vI-\tA$ with $\tA=\vD^{-1/2}\vA\vD^{-1/2}$ where $\vD$ is the degree matrix. 
Next, we briefly introduce the graph signal denoising and fixed point iteration perspective of GNNs that provide a better understanding of the computation trajectory and long-distance dependency in graphs.
Finally, we provide a preliminary study to reveal the computation redundancy in GNNs that motivates the development of LazyGNN.

\subsection{GNNs as Graph Signal Denoising}
\label{sec:gsp}

It is recently established that many popular GNNs can be uniformly understood as solving graph signal denoising problems with various diffusion properties and that the long-distance dependency can be well captured by unrolling various fixed point iterations~\cite{ma2020unified, pan2020_unified,zhu2021interpreting,chen2020graph,gu2020implicit}. For instance, the message passing in GCN~\cite{kipf2016semi}, $\vX_{\text{out}} = \tA \vX_{\text{in}}$, can be considered as one gradient descent iteration for minimizing $\tr(\vX^\top(\vI-\tA)\vX)$ with the initialization $\vX_0=\vX_{\text{in}}$. 
The message passing scheme in APPNP~\cite{klicpera2018predict} follows the aggregation $\vX_{l+1} = (1-\alpha) \tA \vX_l + \alpha \vX_{\text{in}}$ that iteratively minimizes $\|\vX-\vX_\tin\|^2_F + ({1/\alpha}-1)~ \tr (\vX^{\top}  (\vI-\tA) \vX)$ with the initialization $\vX_0=\vX_{\text{in}}$ where $l$ is the index of layers. 
Implicit GNN~\cite{gu2020implicit} adopts projected gradient descent to solve the fixed point problem $\vX=\phi(\vW\vX\tA+\vB)$.
Please refer to the reference~\cite{ma2020unified, pan2020_unified,zhu2021interpreting,chen2020graph} for such understanding of many other popular GNN models. Moreover, a large number of advanced GNN models have been inspired from this perspective~\cite{chen2022optimization,liu2021graph,liu2021elastic,yang2021graph,yang2021implicit,jia2022unifying,jiang2022fmp}.

% \subsection{Preliminary study}
\subsection{Computation Redundancy in Training GNNs}
\label{sec:preliminary}

\begin{figure}[htp]
    \centering
    \subfloat[\centering Without dropout]
    {{\includegraphics[width=3.5cm]{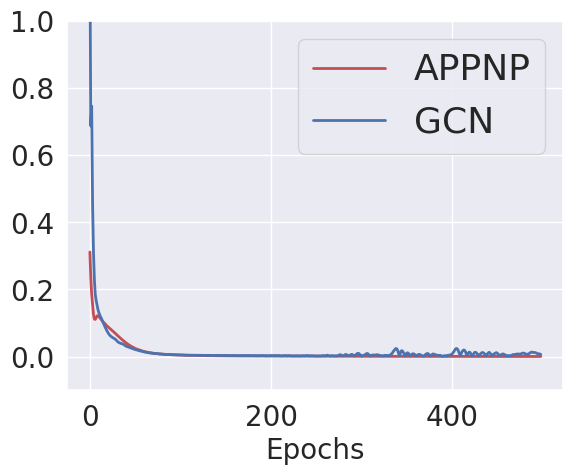}}}
    \hspace{0.2in}
    \subfloat[\centering With dropout]
    {{\includegraphics[width=3.5cm]{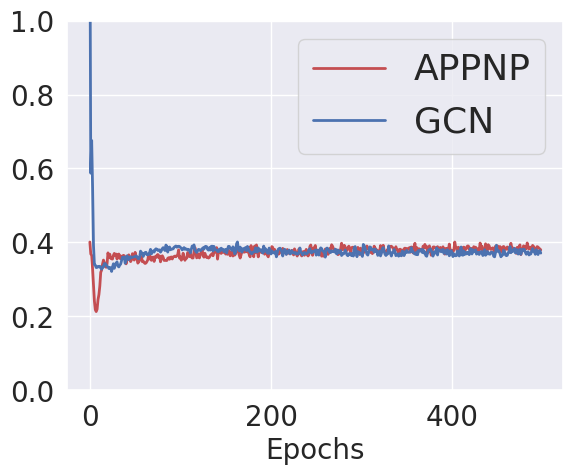}}}%
    % \vspace{-0.1in}
    \caption{Feature changes 
    $\|\vX^{k+1}-\vX^{k}\|_F / \|\vX^{k}\|_F$ on Cora.
    }
    \label{fig:redundancy1}
    % \vspace{-0.1in}
\end{figure}

\begin{figure}[htp]
\vspace{-0.2in}
    \centering
    \subfloat[\centering Without dropout]
    {{\includegraphics[width=3.5cm]{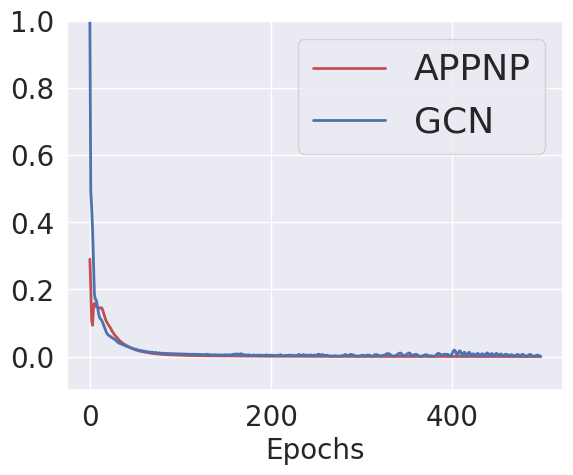}}}
    \hspace{0.2in}
    \subfloat[\centering With dropout]
    {{\includegraphics[width=3.5cm]{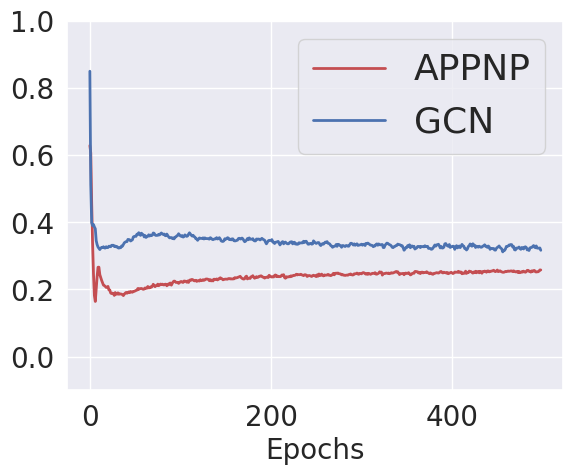}}}%
    % \vspace{-0.1in}
    \caption{Feature changes 
    $\|\vX^{k+1}-\vX^{k}\|_F / \|\vX^{k}\|_F$ on Pubmed.
    }
    \label{fig:redundancy2}
    % \vspace{-0.1in}
\end{figure}

In the training process of GNNs, the model parameters are updated following gradient descent style algorithms such as Adam~\cite{kingma2014adam}. Therefore, the model as well as hidden features in GNNs changes smoothly, especially in the late stages when both the learning rate and gradient norm diminish. This intuition motivates us to investigate the computation redundancy in GNNs. Specifically, we measure the relative changes of the hidden features in the last layer ($L$-th layer) between epochs $k+1$ and $k$, i.e., 
% $\|\vX^{k+1}-\vX^{k}\|_F / \|\vX^{k}\|_F$, 
$\|\vX_L^{k+1}-\vX_L^{k}\|_F / \|\vX_L^{k}\|_F$, 
over the training iterations on Cora and Pubmed dataset ~\cite{kipf2016semi}
using representative models such as GCN~\cite{kipf2016semi} and APPNP~\cite{klicpera2018predict}. We show two cases in Figure~\ref{fig:redundancy1} and Figure~\ref{fig:redundancy2}, depending on whether dropout layers are used. 

The relative changes of hidden features shown in Figure~\ref{fig:redundancy1} and Figure~\ref{fig:redundancy2} demonstrate the following:
(1) when there is no dropout, the hidden features barely change as the training goes; (2) if dropout is used, it will incur additional variation in the hidden features due to the randomness in dropout layers. Both cases demonstrate that the activation values in hidden layers of GNNs indeed change slowly, indicating the existence of \textit{computation redundancy}: the computation in successive training iterations is highly similar. This observation not only validates the rationality of historical embedding used in existing works such as VR-GNN~\cite{chen2017stochastic} and GNNAutoScale~\cite{fey2021gnnautoscale} but also motivates the lazy propagation in this work.

\vspace{-0.2in}

% \vspace{-0.3in}

\section{Lazy Graph Neural Networks}
\label{sec:method}

In this section, we propose a novel shallow LazyGNN that uses a few aggregation layers to capture long-distance dependency in graphs through lazy forward and backward propagations. Then a mini-batch LazyGNN is introduced to handle large-scale graphs with efficient data sampling, computation, and memory usage. 
Complexity analyses are also provided to illustrate the superior scalability of LazyGNN.

\subsection{GNNs with Lazy Propagation}
\label{sec:lazygnn}

Existing research has demonstrated the benefits of capturing long-distance relations in graphs by stacking more feature aggregation layers or unrolling various fixed point iterations as introduced in Section~\ref{sec:intro}. 
However, these deeper GNNs are not efficient due to the neighborhood explosion problem.
Our preliminary study in Section~\ref{sec:preliminary_section} reveals the computation redundancy in GNNs over the training iterations: 
the hidden features in GNNs evolve slowly such that the computation of feature aggregations is highly correlated and redundant over training iterations. 
This observation motivates us to develop 
a novel shallow LazyGNN that captures long-distance relations in graphs by propagating information lazily with very few message-passing layers. 

Without loss of generality, we illustrate the main idea of lazy propagations using the most common and widely used graph signal denoising problem~\cite{zhou2003learning,ma2020unified,yang2021graph}:

\begin{align}
\label{eq:denoise_problem}
\min_{\vX} ~ \|\vX-\vX_\tin\|^2_F + ({1/\alpha}-1)~ \tr (\vX^{\top}  (\vI-\tA) \vX),
\end{align}
\vspace{-0.2in}

\noindent where the first term maintains the proximity with node hidden features $\Xin$ after feature transformation, and the second Laplacian smoothing regularization encodes smoothness assumption on graph representations. 
Note that we only take this as an example to illustrate the main idea of lazy propagation, but the idea can be applied to other GNNs inspired by any denoising problems or fixed point iterations with different properties and design motivations. Next, we illustrate the lazy forward and backward propagations in LazyGNN as well as the innovative designs that solve the technical challenges.

\textbf{Forward Computation.}
From Eq.~\eqref{eq:denoise_problem}, we can derive the high-order iterative graph diffusion as in APPNP~\cite{klicpera2018predict} with $f$ being the feature transformation:
\begin{align}
\vX_\tin^k &= f(\vX_\fea, \Theta^k), \\
\vX_{0}^k &= \vX_\tin^k, \\
\vX_{l+1}^k &= (1-\alpha) \tA \vX_{l}^k+\alpha \vX_\tin^k,~ \forall l=0,\dots,L-1  % \label{eq:mp_iter}
\end{align}
where $l$ and $k$ denote the index of layers and training iterations, respectively. The key insight of LazyGNN is that 
the approximate solution of Eq.~\eqref{eq:denoise_problem} (i.e., $\vX^k_{L}$) evolves smoothly since
$\vX_\tin^k = f(\vX_\fea, \Theta^k)$ changes smoothly with model parameters $\Theta^k$.
Intuitively, the features have been propagated over the graph many times in previous training iterations so it suffices to propagate features lazily by reusing existing computation.

Formally, we propose LazyGNN to leverage the computation redundancy between training iterations by mixing the diffusion output in iteration $k-1$ (i.e., $\vX_L^{k-1}$) into the initial embedding of the diffusion process in training iteration $k$, namely $\vX_0^{k} = (1-\beta) \vX_L^{k-1} + \beta \vX_\tin^k$. This is because, in practice, dropout is commonly used in deep learning to prevent overfitting, which introduces additional variations in the feature embedding as demonstrated in Figure~\ref{fig:redundancy1} and Figure~\ref{fig:redundancy2}. Therefore, we introduce this momentum correction to compensate for such disturbance by mixing current and history features with hyperparameter $\beta$. In practice, small $\beta$ is favored if the dropout rate is small.
To summarize, the forward computation of LazyGNN works as follows:
\begin{align}
\vX_\tin^k &= f(\vX_\fea, \Theta^k), \\
% \vX_{0}^k &= \vX^{k-1}_L, \\
\vX_{0}^k &= (1-\beta) \vX^{k-1}_L + \beta \vX_\tin^k, \label{eq:forward_momentum} \\
\vX_{l+1}^k &= (1-\alpha) \tA \vX_{l}^k+\alpha \vX_\tin^k, ~\forall l=0,\dots,L-1  \label{eq:mp_iter}
\end{align}
\vspace{-0.1in}

\definecolor{ao}{rgb}{0.0, 0.5, 0.0}
\begin{figure*}[htp]
    \centering
    \includegraphics[width=0.8\textwidth]{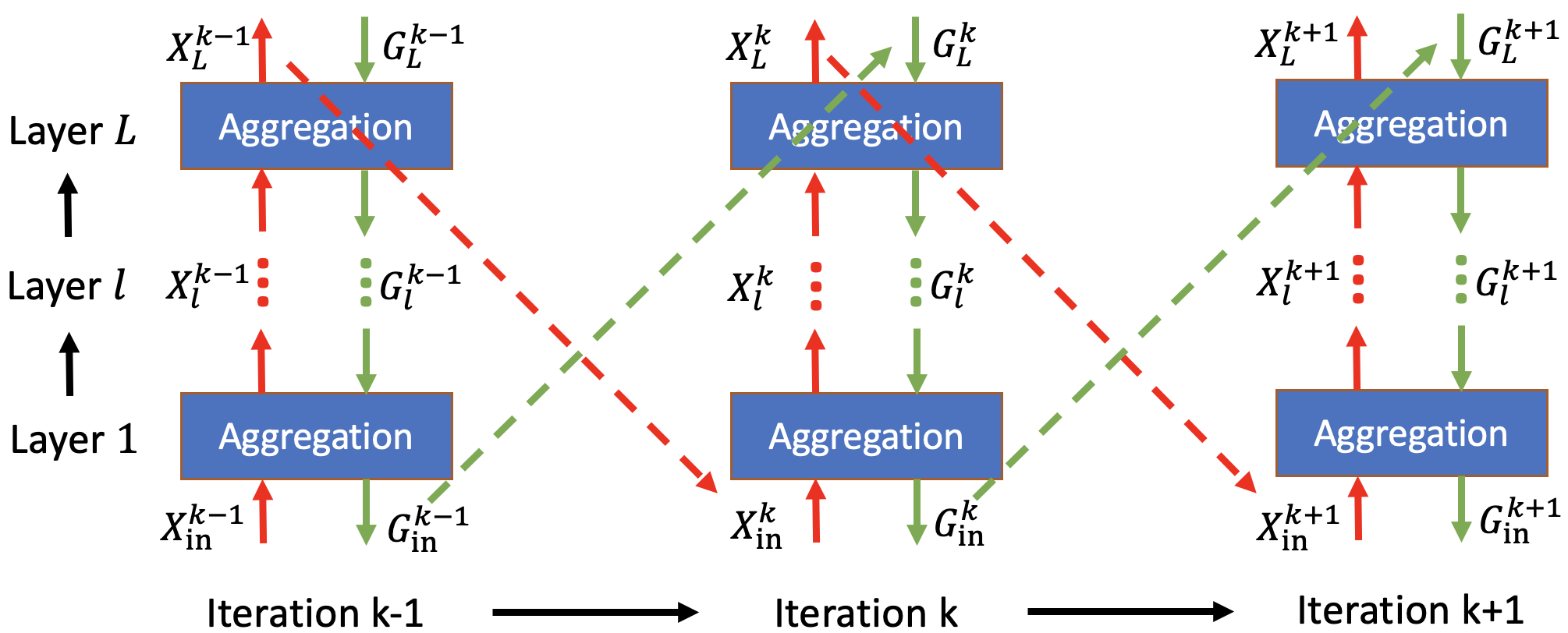}
    % \caption{LazyGNN with Lazy forward (red) and backward (green) propagations.}
    \caption{LazyGNN with Lazy {\color{red}{\textbf{forward}}} (red) and {\color{ao}{\textbf{backward}}} (green) propagations.}
    \label{fig:lazygnn}
\end{figure*}
\vspace{-0.1in}

By lingering the computation over training iterations as shown in Figure~\ref{fig:lazygnn}, the feature aggregation layers in Eq.~\eqref{eq:mp_iter} solve the denoising problem in Eq.~\eqref{eq:denoise_problem} with an implicitly large number of steps although there are only $L$ layers in each training iteration. Therefore, it only requires a few feature aggregation layers (a small $L$) to approximate the fixed point solution $\vX_*^k$ of Eq.~\eqref{eq:denoise_problem}. Note that we find $L=1$ and $L=2$ work very well in our experiments in Section~\ref{sec:experiment}. 
In the extreme case when the learning rate and dropout rate are $0$, the forward computation of LazyGNN is equivalent to running forward propagations $L\times K$ times continuously with $K$ being the total number of training iterations.

% \begin{remark}[Long-distance dependency]
\begin{remark}[\textbf{Long-distance dependency}]
Through lazy propagation, LazyGNN is able to capture long-distance dependency in graphs with a small number of feature aggregation layers because LazyGNN accumulatively propagates information to distant nodes in the graphs over many training iterations. In contrast, existing works try to capture long-distance dependency in graphs by increasing the number of feature propagation layers, which is less efficient.
\end{remark}

% \begin{remark}[Over-smoothing]
\begin{remark}[\textbf{Over-smoothing}]
\label{remark:oversmooth}
In contrast to many GNN models such as GCN or GAT that suffer from the over-moothing issue when more layers are stacked, 
the accumulation of feature aggregations over training iterations in LazyGNN will not cause the over-smoothing issue 
because the residual connection $\vX_\tin^k$ in Eq.~\eqref{eq:mp_iter} determines the fixed point and prevents the over-smoothed solution, as will be verified in Section~\ref{sec:ablation}. 
\end{remark}

\textbf{Backward Computation.}
While it is feasible to leverage the computation from previous training iterations in the forward computation, it is highly non-trivial to compute the gradient for the model update in the backward computation since the computation graphs from previous training iterations have been destroyed and released in the memory. In other words, there is no way to directly compute the backpropagation through the history variables $\{\vX_L^{k-1}, \vX_L^{k-2}, \dots\}$ in current iterations $k$ as being done in sequential models such as RNN~\cite{hochreiter1997long} and Transformer~\cite{vaswani2017attention}. Therefore, we choose to compute the gradient indirectly via the implicit function theorem~\cite{bai2019deep,gu2020implicit}.

\begin{theorem}[Implicit Gradient] 
\label{thm:implicit}
Let $\vX_*$ be the fixed point solution of function $g(\vX_*, \vX_\tin)$, i.e., $g(\vX_*, \vX_\tin)=\vzero$. 
Given the gradient of loss function $\cL(\vX_*, \vY)$ with respect to the fixed point $\vX_*$, i.e., $\frac{\partial \cL}{\partial \vX_*}$, the gradient of loss $\cL$ with respect to feature $\vX_\tin$ can be computed as:
\begin{align}
\frac{\partial \cL}{\partial \vX_\tin} = -\frac{\partial \cL}{\partial \vX_*} (\vJ |_{\vX_*})^{-1} \frac{\partial g(\vX_*, \vX_\tin)}{\partial \vX_\tin} 
\end{align}

where $\vJ|_{\vX_*} = \frac{\partial g(\vX_*, \vX_\tin)}{\partial \vX_*}$ is the Jacobian matrix of $g(\vX_*, \vX_\tin)$ evaluated at $\vX_*$.
\end{theorem}
%
% \vspace{-0.1in}
\begin{proof} 
Take the first-order derivative on both sides of the fixed point equation $g(\vX_*, \vX_\tin)=0$:
\begin{align*}
\frac{ \partial{g(\vX_*, \vX_\tin)}}{\partial \vX_\tin} + \frac{ \partial{g(\vX_*, \vX_\tin)}}{\partial \vX_*} \frac{d \vX_*}{d \vX_\tin} = \vzero 
\end{align*}
\vspace{-0.1in}
\begin{align*}
\frac{d \vX_*}{d \vX_\tin} = - \Big(\frac{ \partial{g(\vX_*, \vX_\tin)}}{\partial \vX_*} \Big )^{-1} \frac{ \partial{g(\vX_*, \vX_\tin)}}{\partial \vX_\tin}
\end{align*}

Using $\frac{\partial \cL}{\partial \vX_\tin} = \frac{\partial \cL}{\partial \vX_*} \frac{d \vX_*}{d \vX_\tin}$, we obtain:
% \vspace{-0.1in}
\begin{align*}
\frac{\partial \cL}{\partial \vX_\tin} = -\frac{\partial \cL}{\partial \vX_*} \Big (\frac{ \partial{g(\vX_*, \vX_\tin)}}{\partial \vX_*} \Big)^{-1} \frac{ \partial{g(\vX_*, \vX_\tin)}}{\partial \vX_\tin}.
\end{align*}
\vspace{-0.1in}
\end{proof}

Specifically, for the problem in Eq.~\eqref{eq:denoise_problem}, we have the fixed point equation: 
\vspace{-0.1in}
% \small{
\begin{align}
\hspace{-0.1in}
g(\vX_*,\vX_\tin)=\vX_*-\vX_\tin + (\frac{1}{\alpha}-1) (\vI-\tA) \vX_* = \vzero
\end{align}
\vspace{-0.2in}

% }
\noindent which gives $\frac{ \partial{g(\vX_*, \vX_\tin)}}{\partial \vX_\tin} = -\vI$ and $\vJ|_{\vX_*}= \frac{1}{\alpha} (\vI-(1-\alpha) \tA)$. 

Therefore, according to Theorem~\ref{thm:implicit}, the gradient of loss $\cL$ with respect to $\Xin$ can be computed as follows:
% \vspace{-0.1in}
\begin{align}
\frac{\partial \cL}{\partial \vX_\tin} = \alpha \frac{\partial \cL}{\partial \vX_*}
\Big ( \vI - (1-\alpha) \tA \Big)^{-1}.
\end{align}
% \vspace{-0.1in}
\vspace{-0.15in}
% \vspace{-0.2in}

\noindent However, it is still infeasible to compute the expensive matrix inversion so we propose to approximate it by the iterative backward gradient propagation: 
% {\small{
% \vspace{-0.1in}
\begin{align} 
\vG_L &= \frac{\partial \cL}{\partial \vX_L} ~~\Big ( \approx \frac{\partial \cL}{\partial \vX_*} \Big ) \\
\vG_{l} &= (1-\alpha) \tA \vG_{l+1} + \alpha \frac{\partial \cL}{\partial \vX_L} ~\forall l=L-1, \dots, 0
\end{align}
% }}
\vspace{-0.2in}

\noindent where $\frac{\partial \cL}{\partial \vX_*}$ is approximated by $\frac{\partial \cL}{\partial \vX_L}$, and $\vG_0$ provides an approximation for gradient $\frac{\partial \cL}{\partial \vX_\tin}$. 
It is clear that the backward computation requires gradient propagation over the graph, which is symmetric to and as expensive as the vanilla forward computation. 
Similarly, to reduce the number of gradient propagation layers, we propose to propagate the gradient lazily by leveraging the gradient $\frac{\partial \cL}{\partial \vX_\tin^{k-1}}$ computed in the previous training iteration $k-1$ as shown in Figure~\ref{fig:lazygnn}:
% \vspace{-0.1in}
% {\small{
% \vspace{-0.1in}
\begin{align}
\vG_L^k &= (1-\gamma) \frac{\partial \cL}{\partial \vX_\tin^{k-1}} + \gamma \frac{\partial \cL}{\partial \vX_L^{k}}  \label{eq:backward_momentum} \\
\vG_{l}^k &= (1-\alpha) \tA \vG_{l+1} + \alpha \frac{\partial \cL}{\partial \vX_L^k} ~\forall l=L-1, \dots, 0
\end{align}
% }}
\vspace{-0.1in}

\noindent where $\frac{\partial \cL}{\partial \vX_*^k}$ is approximated by $\frac{\partial \cL}{\partial \vX_L^k}$, and $\vG_0$ provides an approximation for gradient $\frac{\partial \cL}{\partial \vX_\tin^k}$ so that the gradient of model parameters, i.e., $\frac{\partial \cL}{\partial \Theta^k}$, can be further computed by chain rules.
Similar to the forward computation, the momentum correction in Eq.~\eqref{eq:backward_momentum} compensates the gradient changes over training iterations by mixing the current gradient $\frac{\partial \cL}{\partial \vX_L^{k}}$ and history gradient $\frac{\partial \cL}{\partial \vX_\tin^{k-1}}$ with hyperparameter $\gamma$.

% \begin{remark}[Computation and memory efficiency]
\begin{remark}[\textbf{Computation and memory efficiency}]
Since LazyGNN uses very few aggregation layers, it significantly reduces the computation and memory cost. Moreover, LazyGNN does not need to store intermediate hidden activation values in the aggregation layers because the computation of implicit gradient is agnostic to the forward propagation trajectory, which further reduces memory cost.
\end{remark}

\begin{remark}[\textbf{Communication efficiency}]
LazyGNN provides a fundamental algorithmic improvement to significantly reduce the communication cost in cross-device feature aggregations for distributed GNN training because it uses fewer propagations and communication rounds. This is orthogonal and complementary to many existing system-level strategies that mitigate the significant feature communication overhead in distributed GNN training~\cite{10.5555/3433701.3433794, wan2022pipegcn, wan2022bns, zhang2023boosting}. 
\end{remark}

\subsection{Mini-batch LazyGNN with Subgraph Sampling}
\label{sec:mini-lazygnn}

\begin{figure}[htp]
\vspace{-0.1in}
    \centering
    \includegraphics[width=8.3cm]{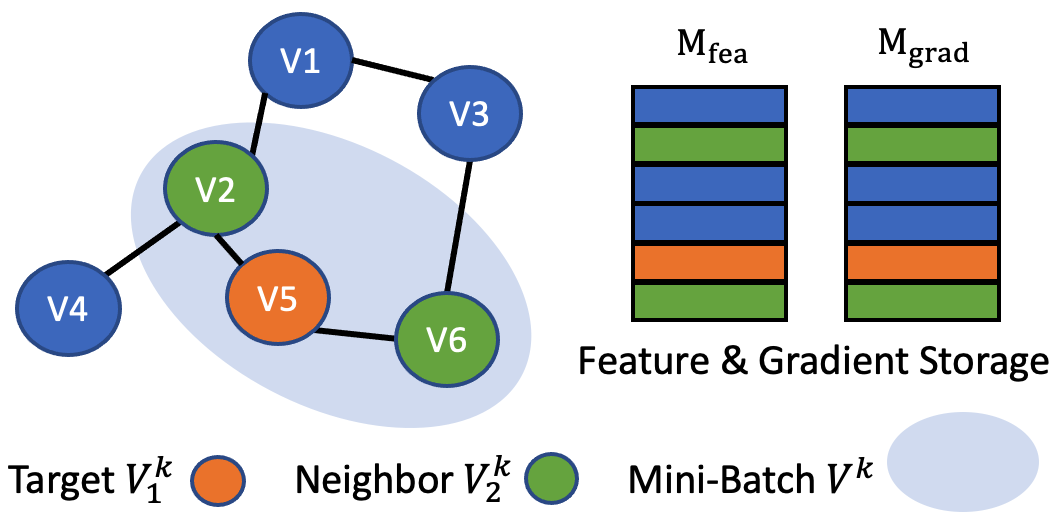}
    % \vspace{-0.1in}
    \caption{Mini-batch LazyGNN with Feature \& Gradient Storage.}
    \label{fig:mini-batch}
\end{figure}

As illustrated in Section~\ref{sec:lazygnn}, LazyGNN solves the inherent neighborhood explosion problem so that when handling large-scale graphs, the mini-batch sampling only needs to sample neighboring nodes within a few hop distances. To further demonstrate the superior scalability, we introduce a mini-batch LazyGNN that enhances the computation and memory efficiency with efficient data sampling as shown in Figure~\ref{fig:mini-batch}.
In each training iteration $k$, we sample a target node set $V_1^k$ and their $L$-hop neighbor set $V_2^k$, and we denote the union of these two nodes as $V^k=V_1^k \cup V_2^k$. An induced subgraph $\tA_{V^k}$ is constructed based on the node set $V^k$. Note that LazyGNN works well with small $L\in\{1,2\}$ so that the data sampling is extremely efficient.

\textbf{Forward Computation.}
The forward computation of mini-batch LazyGNN works as follows:
% \vspace{-0.1in}
\begin{align}
(\vX_\tin^k)_{V_k} &= f \big((\vX_\fea)_{V^k}, \Theta^k \big), \\
(\vX_0)_{V_k} &= (1-\beta) (\vM_\fea)_{V^k} 
+ \beta (\vX_\tin^k)_{V_k} \label{eq:mini_mix_forward}\\ %~~(\text{history}) \\ 
(\vX_{l+1}^k)_{V^k} &= (1-\alpha) \tA_{V^k} (\vX_{l}^k)_{V^k} + \alpha (\vX_\tin^k)_{V^k}, 
\forall l \label{eq:mini_forward_prop} \\
% \forall l =0,\dots,L-1  \\
(\vM_\fea)_{V^k_1} &= (\vX_{L}^k)_{V^k_1}  \label{eq:save_forward_history}
\end{align}
% \vspace{-0.2in}
% \vspace{-0.3in}
\vspace{-0.1in}

\noindent The node features $(\vX_\fea)_{V^k}$ for the node set $V^k$ are sampled to form a mini-batch. The corresponding diffused node features $(\vM_\fea)_{V^k}$ of the same node set are retrieved from the feature storage $\vM_\fea$ and then combined with current node features $(\vX_\tin^k)_{V_k}$ in Eq.~\eqref{eq:mini_mix_forward}. The lazy forward propagation runs on the subgraph $\tA_{V^k}$ in Eq.~\eqref{eq:mini_forward_prop}. Finally, $(\vX_{L}^k)_{V^k_1}$ provides the prediction for target nodes $V_1^k$, which is further maintained in the features storage $\vM_\fea$.
The embeddings of neighbor nodes $(\vX_{L}^k)_{V^k_2}$ are not accurate and not stored.
% since their computation is not accurate. 

\textbf{Backward Computation.}
The backward computation of mini-batch LazyGNN works as follows:
% \vspace{-0.1in}
% {\small{
\begin{align}
& (\vG_L^k)_{V^k} = (1-\gamma) (\vM_\grad)_{V^k} + \gamma \frac{\partial \cL}{\partial (\vX_L^{k})_{V^k} }   \label{eq:mini_mix_backward} \\
& (\vG_{l}^k)_{V^k} = (1-\alpha) \tA_{V^k} (\vG_{l+1})_{V^k} + \frac{ \alpha ~~\partial \cL}{\partial (\vX_L^k)_{V^k} } ~\forall l \label{eq:mini_backward_prop} \\ %~\forall l=L-1, \dots, 0
 & (\vM_\grad)_{V^k_1} = (\vG_{0}^k)_{V^k_1} \label{eq:mini_backward_history}
\end{align}
% }}
\vspace{-0.2in}

\noindent In Eq.~\eqref{eq:mini_mix_backward}, $(\vM_\grad)_{V^k}$, the history gradient with respect to node features $(\vX_{\tin}^{k-1})_{V^k}$, is retrieved from the gradient storage $\vM_\grad$ and then combined with $\frac{\partial \cL}{\partial (\vX_L^{k})_{V^k}}$, the gradient with respect to current node features $(\vX_L^{k})_{V^k}$. The lazy backward propagation runs on the subgraph $\tA_{V^k}$ in Eq.~\eqref{eq:mini_backward_prop}. Finally, $(\vG_{0}^k)_{V^k_1}$ provides an estimation for $\frac{\partial \cL}{\partial (\vX_\tin^{k})_{V^k_1} }$, which is used to compute gradient of parameters $\Theta^k$ by chain rules and then maintained in the gradient storage $\vM_\grad$. 
Likewise, the gradients of neighbor nodes $\frac{\partial \cL}{\partial (\vX_\tin^{k})_{V^k_2} }$ are not accurate and will not be stored in the gradient memory.

\subsection{Complexity Analysis}

As discussed in Section~\ref{sec:lazygnn} and Section~\ref{sec:mini-lazygnn}, LazyGNN is scalable due to its efficiency in computation, memory, data sampling, and communication.
% , and potential parallelism. 
In this section, we provide complexity analyses to further demonstrate its superior scalability. Since the major efficiency difference lies in feature aggregation layers, we focus on the complexity of feature aggregations.
Suppose $L$ is the number of propagation layers, $H$ is the size of hidden units,
$N$ is the number of nodes, $M$ is the number of edges. For simplicity, we assume that $H$ is the same for all hidden layers.

\textbf{Computation complexity.} 
Performing one feature aggregation requires a sparse-dense matrix multiplication that needs $\cO(MH)$ operations.
Therefore, the computation complexity for forward feature aggregations and backward gradient aggregations is $\cO(2LMH)$ per epoch. However, the number of layers $L$ in LazyGNN is much smaller than existing approaches, so it significantly reduces the computation cost.

\textbf{Memory complexity.} 
% For memory complexity, 
$\cO(NH)$ memory space is required for the storage of history feature $\vX_L^k$ (i.e., $\vM_\fea$) and history gradient $\frac{\partial \cL}{\partial \vX_\tin^k}$ (i.e., $\vM_\grad$) in LazyGNN. LazyGNN does not store the intermediate state at each feature aggregation layer because the backward gradient computation is agnostic to the forward computation trajectory. Therefore, the total memory complexity for LazyGNN is $\cO(NH)$, which is independent of the number of aggregation layers. In contrast, existing state-of-art history-embedding based models, such as GNNAutoScale~\cite{fey2021gnnautoscale} and GraphFM~\cite{yu2022graphfm}, require the storage of feature embeddings in each layer, which leads to the memory complexity $\cO(LNH)$. Moreover, they also require the storage of all intermediate layers for gradient computation. Their memory cost increases linearly with the
number of layers, which is essential in capturing long-distance dependency in those models. In fact, because of the large storage requirements, GAS chooses to save the history embedding in CPU memory, but the data movement between CPU and GPU can be dominating compared with the movement in GPU as verified in Section~\ref{sec:efficiency}. 

\textbf{Sampling efficiency.}
Mini-batch LazyGNN further reduces the computation and memory cost by data sampling, which enjoys the same benefits as existing sampling-based methods. Compared to classic sampling-based models, such as GraphSAGE and GraphSAINT, mini-batch LazyGNN only needs to sample neighbors in one hop or two hops distance so that the data sampling is efficient. This data sampling efficiency might bring significant practical speedup for large-scale problems where the data loading of node features and subgraphs can be dominating~\cite{ma2022bifeat}.

\textbf{Communication efficiency.}
Recent works~\cite{10.5555/3433701.3433794, wan2022pipegcn, wan2022bns, zhang2023boosting} clearly demonstrate that the significant feature communication overhead in propagation layers dominates the running time in the parallel training of GNNs. Specifically, the communication time divided by the total training time is over 80\% for ogbn-products as shown in the work~\cite{wan2022pipegcn}. Therefore, existing works propose various system-level strategies to mitigate this bottleneck.
Although this work does not focus on the distributed setting, it is worth emphasizing that LazyGNN is promising for parallelism since it requires fewer communication rounds due to lazy propagation.

Above analyses indicate that LazyGNN is highly efficient in computation, memory, data sampling, and communication. The practical running time and memory cost will be discussed in Section~\ref{sec:efficiency}.

% \vspace{-0.1in}
\section{Experiments}
\label{sec:experiment}

In this section, we provide comprehensive experiments to validate the superior prediction performance and scalability of LazyGNN. Specifically, we try to answer the following questions: 
(Q1) Can LazyGNN achieve strong prediction accuracy on large-scale graph benchmarks? (Section~\ref{sec:performance})
(Q2) Can LazyGNN handle large graphs more efficiently than existing approaches? (Section~\ref{sec:efficiency})
(Q3) What are the impacts of the hyperparameters in LazyGNN? (Section~\ref{sec:ablation})

\subsection{Prediction Performance}
\label{sec:performance}

\textbf{Experimental settings.} 
We conduct experiments on multiple large-scale graph datasets including REDDIT, YELP, FLICKR, ogbn-arxiv, and ogbn-products~\cite{hu2020open}. We evaluate the graph representation learning by node classification accuracy in the semi-supervised setting. We provide a performance comparison with multiple baselines including GCN~\cite{kipf2016semi}, GraphSage~\cite{hamilton2017inductive}, FastGCN~\cite{chen2018fastgcn}, LADIES~\cite{zou2019layer}, VR-GCN~\cite{chen2017stochastic}, MVS-GNN~\cite{cong2020minimal}, Cluster- GCN~\cite{chiang2019cluster}, GraphSAINT~\cite{Zeng2020GraphSAINT}, SGC~\cite{wu2019simplifying}, SIGN~\cite{rossi2020sign}, GAS~\cite{fey2021gnnautoscale} and VQ-GNN~\cite{ding2021vq}. 
The hyperparameter tuning of baselines closely follows the setting in GNNAutoScale~\cite{fey2021gnnautoscale}. 

For LazyGNN, hyperparameters are tuned from the following search space: (1) learning rate: $\{0.01, 0.001, 0.0001\}$;  (2) weight decay: $\{0, 5e-4, 5e-5\}$; (3) dropout: $\{0.1, 0.3, 0.5, 0.7\}$; (4) propagation layers : $L \in \{1, 2\}$; (5) MLP layers: $\{3,4\}$; (6) MLP hidden units: $\{256, 512\}$; (7) $\alpha \in \{0.01, 0.1, 0.2, 0.5, 0.8\}$; (8) $\beta$ and $\gamma$ are simply set as $0.5$ in most cases, but a further tuning can improve the performance.
% For data sampling, w
While LazyGNN is compatible with any sampling method, we adopt the subgraph sampling strategy as in GNNAutoScale to ensure a fair comparison.

\textbf{Performance analysis.} From the accuracy as summarized in Table~\ref{tab:prediction_acc}, we can make the following observations:
\vspace{-0.1in}
\begin{itemize}[leftmargin=0.15in]
\item 
% \textbf{(1)} 
LazyGNN achieves state-of-art performance on almost all datasets. In particular, LazyGNN achieves $73.2\%$ and $82.3\%$ accuracy on ogbn-arxiv and ogbn-products. 
The only exceptions are that GraphSAINT slightly outperforms LazyGNN by $0.2\%$ on REDDIT and that GCNII outperforms LazyGNN by $1.1\%$ on FLICKR. However, LazyGNN is much more efficient than GraphSAINT and GCNII in computation and memory cost. Note that GCNII achieves the reported performance with 10 transformation layers and 8 propagation layers while LazyGNN only uses 4-layer MLP and 2-layer propagation. We believe a stronger performance can be achieved with a more thorough architecture tuning on LazyGNN.

\item 
% \textbf{(2)} 
Importantly, we observe that full-batch LazyGNN closely matches mini-batch LazyGNN, which indidates that the subgraph sampling in LazyGNN can significantly improve the scalibility (as will be discussed in Section~\ref{sec:efficiency}) but do not heavily hurt the prediction performance. Sometimes, LazyGNN even improves the generalization performance due to sampling (such as for REDDIT).

\item 
% (3) 
LazyGNN also consistently outperforms APPNP with 10 propagation layers, which indicates the advantage of capturing long-distance dependency in graphs by lazy propagation. The reproduction of GAS+GCN on YELP is much worse than reported in the work~\cite{fey2021gnnautoscale}, therefore we omit the result.
\end{itemize}

\begin{table}[ht!]
\caption{Prediction accuracy ($\%$) on large-scale graph datasets. ``full'' and ``mini'' stand for full-batch and mini-batch respectively. OOM stands for out-of-memory.}
\label{tab:prediction_acc}
% \vskip 0.15in
\vspace{-0.1in}
\begin{center}
\begin{small}
% \begin{sc}
\setlength{\tabcolsep}{2pt}
\resizebox{\linewidth}{!}{%
\begin{tabular}{lcccccc}
\toprule
    \mc{1}{l}{\footnotesize{\textbf{\#\,nodes}}} & \footnotesize{230K} & \footnotesize{89K} & \footnotesize{717K} & \footnotesize{169K} & \footnotesize{2.4M} \\ [-0.1cm]
    \mc{1}{l}{\footnotesize{\textbf{\#\,edges}}} & \footnotesize{11.6M} & \footnotesize{450K} & \footnotesize{7.9M} & \footnotesize{1.2M} & \footnotesize{61.9M} \\ [-0.05cm]
    \mc{1}{l}{\mr{2}{\textbf{Method}}} & \mr{2}{\textsc{Reddit}} & \mr{2}{\textsc{Flickr}} & \mr{2}{\textsc{Yelp}} & \texttt{ogbn-} & \texttt{ogbn-} \\
    & & & & \texttt{arxiv} & \texttt{products} \\
    % \midrule
\midrule
GraphSAGE    & 95.4 & 50.1 & 63.4 & 71.5 & 78.7 \\
FastGCN    & 93.7 & 50.4 & --- & --- & ---  \\
LADIES    & 92.8 & --- & --- & --- & --- \\
VR-GCN    & 94.5 & --- & 61.5 & --- & ---  \\
MVS-GNN    & 94.9 & --- & 62.0 & --- & ---  \\
Cluster-GCN    & 96.6 & 48.1 & 60.9 & --- & 79.0  \\
GraphSAINT    & \textbf{97.0} & 51.1 & 65.3 & --- & 79.1  \\
SGC    & 96.4 & 48.2 & 64.0 & --- & ---  \\
SIGN    & 96.8 & 51.4 & 63.1 & --- & 77.6  \\
VQ-GNN    & 94.5 & --- & --- & 70.6 & ---  \\
\hline
GCN (full)    & 95.4 & 53.7 & OOM & 71.6 & OOM   \\
APPNP (full)  & 96.1 & 53.4 & OOM & 71.8 & OOM\\
GCNII (full)    & 96.1 & 55.3 & OOM & 72.8 & OOM    \\
\hline
GCN (GAS)   & 95.4 & 54.0 & --- & 71.7 & 76.7   \\
APPNP (GAS)  & 96.0 & 52.4 & 63.8 & 71.9 &76.2 \\
GCNII (GAS)   & 96.7 & \textbf{55.3} & 65.1 & 72.5 & 77.2   \\
\hline
LazyGNN (full)   & 96.2& 54.2 & OOM & \textbf{73.2} & OOM   \\
LazyGNN (mini)   & 96.8& 54.0 & \textbf{65.4}& 73.0 & \textbf{82.3}    \\
\bottomrule
\end{tabular}
}
% \end{sc}
\end{small}
\end{center}
% \vskip -0.1in
\vspace{-0.1in}
% \end{table*}
\end{table}

\subsection{Efficiency Analysis}
\label{sec:efficiency}

To verify the scalability of LazyGNN, we provide empirical efficient analysis compared with state-of-art end-to-end trainable and scalable GNNs such as GNNAutoScale (GAS) since it has been shown to be the most efficient algorithm for training large-scale GNNs~\cite{fey2021gnnautoscale}.
Specifically, we measure the memory usage and running time for the training on ogbn-products dataset using the same batch size. 
We run GAS using the authors' code which stores the feature embedding for each layer in CPU memory by default since it requires large memory for embedding storage. We also re-implement GAS by storing all memory on the GPU for a better comparison. For LazyGNN, the memory for storing two small tensors (i.e., feature and gradient) is required. Furthermore, we measure the running time of all models for two cases depending on whether the storage is on CPU or GPU memory. Note that all methods use similar number of training epochs as will be verified in Section~\ref{sec:ablation}, so we compare them using the running time per epoch.
For GAS baselines, we use the architecture that can reach the best performance.
From the measurements summarized in Table~\ref{tab:memory}, we can make the following observations: 
\vspace{-0.1in}
\begin{itemize}[leftmargin=0.15in]
\item LazyGNN (GPU) has a much shorter running time (7.7s) per epoch compared with all baselines. All methods become much slower if the memory needs to be moved between GPU and CPU.

\item LazyGNN only requires small memory space and  
% memory efficient since 
only 878 MB of additional storage is required to store two small tensors for feature and gradient regardless of the number of layers. 
In contrast, GAS requires large memory storage due to the feature storage for each layer. This verifies that LazyGNN is memory efficient.
% and computation.

\item LazyGNN (GPU+CPU) is faster than APPNP (GAS) because APPNP requires more data movement for feature propagation. This further verifies the computation and memory efficiency of LazyGNN. 

An even better running time of LazyGNN is expected if the data movement is optimized as the implementation in GAS using asynchronous calls, but we leave it as a future work.
\end{itemize}

\vspace{-0.1in}
% \begin{table*}[ht!]
\begin{table}[ht!]
\caption{Memory usage (MB) and running time (seconds per epoch) on ogbn-products.}
\label{tab:memory}
\vskip 0.1in
\begin{center}
% \begin{small}
% \begin{sc}
\setlength{\tabcolsep}{2pt}
\resizebox{1.0\linewidth}{!}{%
\begin{tabular}{lcccccc}
\toprule
\mc{1}{l}{\mr{2}{\textbf{Method}}} & \mr{2}{\textbf{Setting}} & \textsc{GPU} & \textsc{CPU} & \textsc{Running} \\
& & {\textsc{Memory}} & {\textsc{Memory}} & {\textsc{Time}} \\
\midrule
GCN (GAS) & \mr{4}{GPU+CPU} & 2512  & 4783 & 28.8\\
GCNII (GAS) & & 3035 & 4783 & 44.3\\
APPNP (GAS) & & 2142  & 3952 & 84.5\\
LazyGNN (mini) & & 2101 & 878 & 44.5\\
\hline
GraphSage & \mr{5}{GPU Only}  & 4781  & --- & 33.7\\
GCN (GAS) & & 7296 & --- & 11.7\\
GCNII (GAS) & & 7820 & --- & 18.2\\
APPNP (GAS) & & 6097 & --- & 10.5\\
LazyGNN (mini) & & \textbf{2981} & --- & \textbf{7.7} \\
\bottomrule
\end{tabular}
}
\end{center}
% \end{table*}
\vspace{-0.1in}
\end{table}

\subsection{Ablation Study}
\label{sec:ablation}
We provide detailed ablation studies on the impacts of hyperparameter settings in LazyGNN.

\textbf{Oversmoothing.} 
We measure the L2 distance between connected nodes to show the smoothness of GCN, APPNP, and LazyGNN on Cora. Table~\ref{tab:oversmooth} shows that the smoothness value of GCN becomes smaller and smaller as the number of layers increases, which means the node features become over-smoothed and indistinguishable. But for LazyGNN and APPNP, when more layers are being stacked, the smoothness values become stable so that the node features can still be distinguished without suffering from over-smoothing. This provides direct evidence that LazyGNN does not suffer from the over-smoothing problem as discussed in Remark~\ref{remark:oversmooth}.

\vspace{-0.1in}
\begin{table}[ht!]
\caption{$L_2$ feature distance between connected nodes (smoothness) with a varying number of propagation layers on Cora.}
\label{tab:oversmooth}
% \vskip 0.15in
\begin{center}
% \begin{small}
% \begin{sc}
\resizebox{0.7\linewidth}{!}{%
\begin{tabular}{lccccc}
\toprule
{\textbf{Layers}} & \textsc{GCN} & \textsc{APPNP} & \textsc{LazyGNN} \\
\midrule
L = 1  & 336  & 353 & 302\\
L = 5  & 315  & 305 & 285\\
L = 10 & 297 & 296 & 290\\
L = 15 & 263  & 304 & 291\\
L = 20 & 225 & 302 & 292\\
\bottomrule
\end{tabular}
}
% \end{sc}
% \end{small}
\end{center}
\vskip -0.1in
\end{table}

\textbf{Lazy Propagation.} 
We conduct the ablation study to analyze the impact of feature propagation layers and long-distance dependency. Specifically, we use the same MLP for LazyGNN and APPNP and change the number of propagation layers $L$. The comparison in Table~\ref{tab:layers} shows that: 

\vspace{-0.2in}
\begin{itemize}[leftmargin=0.15in]
\item 
The performance of APPNP (GAS) improves with the use of more propagation layers (larger $L$). This trend is pretty clear on ogbn-arxiv, which verifies the benefits of capturing long-distance dependency on graphs. The improvement on ogbn-products is relatively minor due to the large staleness of feature memory caused by GAS when a smaller batch size has to be used.

\item 
% (2) 
LazyGNN archives great performance with even $1$ propagation layer ($72.5\%$ for ogbn-arxiv and $79.8\%$ for ogbn-products). It has already achieved state-of-the-art performance ($73.0\%$ for ogbn-arxiv and $82.3\%$ for ogbn-products) with $2$ propagation layers. These results confirm LazyGNN's advantages and capability in capturing long-distance dependency in graphs with very few feature propagation layers.
\end{itemize}

\vspace{-0.2in}
\begin{table}[ht!]
\caption{Prediction accuracy with a varying number of propagation layers on ogbn-arxiv and ogbn-products.}
\vskip-0.1in
\label{tab:layers}
\begin{center}
% \begin{small}
% \begin{sc}
\resizebox{1.0\linewidth}{!}{%
\begin{tabular}{lcc}
\toprule
\textbf{Method} & \textsc{ogbn-arxiv (\%)} & \textsc{ogbn-products (\%)}  \\
\midrule
APPNP (GAS) (L=1) &70.6 & 75.2\\
APPNP (GAS) (L=2) & 71.4 & 75.5\\
APPNP (GAS) (L=3) &71.5 & 75.7\\
APPNP (GAS) (L=4) &71.8 & 75.8\\
APPNP (GAS) (L=5) &72.2 & 76.0\\
APPNP (GAS) (L=8) &72.5 & 76.1\\
APPNP (GAS) (L=10) &72.7 & 76.2\\
\hline
LazyGNN (L=1)   & 72.5 & 79.8\\
LazyGNN (L=2)   & \textbf{73.0} & \textbf{82.3} \\
LazyGNN (L=3)   & \textbf{73.0} & OOM \\
\bottomrule
\end{tabular}
}
% \end{sc}
% \end{small}
\end{center}
\vskip -0.1in
\end{table}

\textbf{Batch Size.} 
We investigate the impact of the batch size on data sampling. 
Note that we adopt a cluster-based subgraph sampling so that the batch size is determined by the number of clusters since one cluster is sampled in each batch. In other words, more clusters result in a smaller batch size.

The comparison in Table~\ref{tab:batch} demonstrates that a larger batch size (less clusters) brings better performance because the sampling variance is reduced. Overall the performance is quite stable using varying batch sizes (clusters).

\vspace{-0.1in}
\begin{table}[ht!]
\caption{Prediction accuracy of mini-batch LazyGNN with different batch sizes on ogbn-products.}
\label{tab:batch}
% \vskip 0.15in
\begin{center}
% \begin{small}
% \begin{sc}
\resizebox{0.7\linewidth}{!}{
\begin{tabular}{lcccc}
\toprule
\textbf{Clusters} & \textsc{50} & \textsc{100} & \textsc{150} & \textsc{200}\\
\midrule
Accuracy(\%) & 82.3 & 81.8 & 81.5 & 80.6\\

\bottomrule
\end{tabular}
}
% \end{sc}
% \end{small}
\end{center}
\vskip -0.1in
\end{table}

\textbf{Sensitivity Analysis.} 
We provide a detailed sensitivity analysis for the two additional hyperparameters $\beta$ and $\gamma$ in LazyGNN. 
The results in Table~\ref{tab:hyperparameter-ablation} show that: (1) $\{\beta=0, \gamma=0\}$ produces the worst performance ($71.0\%$) because it fully trusts the history embedding that might be outdated due to the additional variations caused by dropout as demonstrated in Section~\ref{sec:preliminary}; 
(2) $\{\beta=1, \gamma=1\}$ performs slightly better ($71.7\%$) even without lazy propagation because 2 propagation layers can provide a good approximate solution;
(3) $\{\beta=0.5, \gamma=0.5\}$ performs the best ($73.0\%$) because it achieves a good trade-off between historical information and current iteration. It compensates for the staleness in the history storage as designed;
(4) The performance of LazyGNN is quite stable for a large range setting of $\beta$ and $\gamma$. In fact, we fix $\beta=\gamma=0.5$ for LazyGNN in most experiments for simplicity. Therefore, it requires almost negligible effort for additional hyperparameter tuning.

\vspace{-0.1in}
\begin{table}[ht!]
\caption{Prediction accuracy of mini-batch LazyGNN for different hyperparameter settings on ogbn-arxiv.}
\label{tab:hyperparameter-ablation}
% \vskip 0.1in
\begin{center}
% \begin{small}
% \begin{sc}
% \begin{tabular}{lc}
\begin{tabular}{cc}
\toprule
% LazyGNN Variants & Accuracy (\%) \\
\textbf{Hyperparameter settings} & \textsc{Accuracy (\%)} \\
\midrule
$\beta=0.0,~\gamma=0.0$ & 71.0\\
$\beta=1.0,~\gamma=1.0$ & 71.7\\
\midrule
$\beta=0.5,~\gamma = 0.1$  & 72.5\\
$\beta=0.5,~\gamma = 0.5$  & 73.0 \\
$\beta=0.5,~\gamma = 0.9$  &  72.8\\
\midrule
$\beta=0.1,~\gamma=0.5$  & 72.4\\
$\beta=0.5,~\gamma=0.5$  & 73.0 \\
$\beta=0.9,~\gamma=0.5$  & 72.7\\
\bottomrule
\end{tabular}
% \end{sc}
% \end{small}
\end{center}
\vskip -0.1in
\end{table}

\begin{figure}[htp]
    \centering
    \includegraphics[width=6cm]{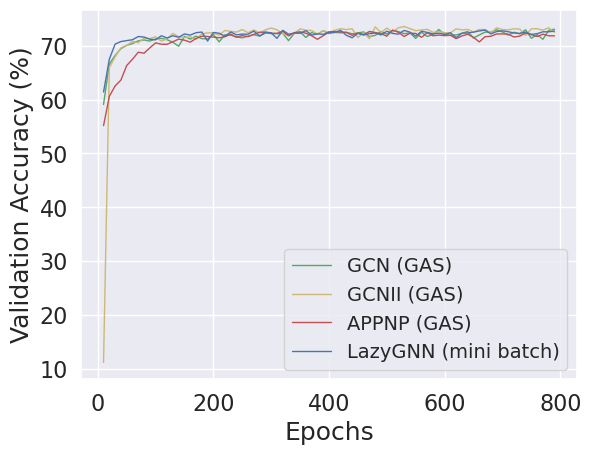}
    \vspace{-0.1in}
    \caption{Validation performance versus training epochs.}
    \label{fig:convergence}
    % \vspace{-0.1in}
\end{figure}

\textbf{Convergence.} 
We show the convergence of LazyGNN during the training process using ogbn-arxiv dataset. 
The convergence of validation accuracy in Figure~\ref{fig:convergence} demonstrates that LazyGNN has a comparable convergence speed with GCN (GAS) and GCNII (GAS), and is slightly faster than APPNP (GAS) in terms of the number of training epochs.

% \vspace{-0.1in}

%%%%%%%%%%%%%%%%%%%%%%%%%%%%%%%%%%%%%%%%%%%%%%%%%%%%%%%%%%%%%%%%%%%%%%%%%%%%%%%%%%%%%%%%%%%%%%%%%%%%%%%%%%%%%%%%%%%%%%%%%%%%%%%%%%%%%%%%%%%%%%%%%%%%%%%%%%%%

\section{Related Work}

It has been generally demonstrated that it is beneficial to capture long-distance relations in graphs by stacking more feature aggregation layers or unrolling various fixed point iterations in GNNs~\cite{gasteiger2018predict, gu2020implicit, liu2020towards, chen2020simple, li2021training, ma2020unified, pan2020_unified, zhu2021interpreting, chen2020graph}. But these works suffer from scalability concerns due to the neighborhood explosion problem~\cite{hamilton2017inductive}. 

A large body of existing research focuses on improving the efficiency and scalability of large-scale GNNs using various novel designs, such as sampling methods, pre-computing or post-computing methods, and distributed methods.
% ========
Sampling methods adopt mini-batch training strategies to reduce computation and memory requirements by sampling nodes and edges. They mitigate the neighbor explosion issue by either removing neighbors~\cite{hamilton2017inductive, chen2018fastgcn, Zeng2020GraphSAINT, zou2019layer} or updating with feature memory~\cite{fey2021gnnautoscale, yu2022graphfm}. 
Pre-computing or post-computing methods separate the feature aggregation and prediction models into two stages, such as pre-computing the feature aggregation before training~\cite{wu2019simplifying, rossi2020sign, sun2021scalable, zhang2022graph, bojchevski2020scaling} or post-processing with label propagation after training~\cite{zhu2005semi, huang2020combining}. 
Distributed methods distribute large graphs to multiple servers and parallelize GNNs training 
~\cite{chiang2019cluster, chai2022distributed, shao2022distributed}. 
In retrospect, these existing approaches
still suffer from various limitations, such as high costs in computation, memory, and communication as well as performance degradation due to large approximation errors or multi-stage training.

Different from existing works, in this work, we propose LazyGNN from a substantially different and novel perspective and propose to capture the long-distance dependency in graphs by shallower models instead of deeper models. This leads to a much more efficient LazyGNN for graph representation learning. Moreover, existing approaches for scalable GNNs can be used to further accelerate LazyGNN. The proposed mini-batch LazyGNN is a promising example.

LazyGNN also draws insight from existing research on graph signal processing and implicit modeling. The forward construction of LazyGNN follows the optimization perspective of GNNs~\cite{ma2020unified}, and it is can be easily generalized to other fixed point iterations with various diffusion properties. 
LazyGNN is related to recent works in implicit modeling, such as Neural Ordinary Differential Equations~\cite{chen2018neural}, Implict Deep Learning~\cite{el2021implicit}, Deep Equilibrium Models~\cite{bai2019deep}, and Implicit GNNs~\citep{gu2020implicit}. 
LazyGNN focuses on developing shallow models with highly scalable and efficient computation while these implicit models demand significantly more computation resources.

\section{Conclusions}

We propose LazyGNN, a novel shallow model to solve the neighborhood explosion problem in large-scale GNNs while capturing long-distance dependency through lazy propagation. We also develop a highly scalable and efficient variant, mini-batch LazyGNN, to handle large graphs. Comprehensive experiments demonstrate its superior prediction performance and efficiency on large-scale problems. The proposed lazy propagation provides a promising algorithmic strategy complementary to existing efforts. We plan to explore its application on other types of GNN models and its further acceleration using other scalable techniques in the future.


\bibliography{ref}
\bibliographystyle{icml2023}

%%%%%%%%%%%%%%%%%%%%%%%%%%%%%%%%%%%%%%%%%%%%%%%%%%%%%%%%%%%%%%%%%%%%%%%%%%%%%%%
%%%%%%%%%%%%%%%%%%%%%%%%%%%%%%%%%%%%%%%%%%%%%%%%%%%%%%%%%%%%%%%%%%%%%%%%%%%%%%%
% APPENDIX
%%%%%%%%%%%%%%%%%%%%%%%%%%%%%%%%%%%%%%%%%%%%%%%%%%%%%%%%%%%%%%%%%%%%%%%%%%%%%%%
%%%%%%%%%%%%%%%%%%%%%%%%%%%%%%%%%%%%%%%%%%%%%%%%%%%%%%%%%%%%%%%%%%%%%%%%%%%%%%%

% \input{section/appendix}

%%%%%%%%%%%%%%%%%%%%%%%%%%%%%%%%%%%%%%%%%%%%%%%%%%%%%%%%%%%%%%%%%%%%%%%%%%%%%%%
%%%%%%%%%%%%%%%%%%%%%%%%%%%%%%%%%%%%%%%%%%%%%%%%%%%%%%%%%%%%%%%%%%%%%%%%%%%%%%%

\end{document}